\documentclass[letterpaper, 10 pt, conference]{ieeeconf}
\IEEEoverridecommandlockouts
\usepackage{cite}
\usepackage{amsmath,amssymb,amsfonts}
\usepackage{algorithmic}
\usepackage{graphicx}
\usepackage{textcomp}
\usepackage{xcolor}
\usepackage[T1]{fontenc} 
\usepackage{amsmath}
\usepackage{mdframed}
\usepackage{comment}
\usepackage[cmintegrals]{newtxmath}
\usepackage{bm} 
\usepackage{subcaption}
\def\BibTeX{{\rm B\kern-.05em{\sc i\kern-.025em b}\kern-.08em
    T\kern-.1667em\lower.7ex\hbox{E}\kern-.125emX}}

\DeclareMathOperator*{\argmin}{arg\,min}

\newcommand{\ambspace}{X}
\newcommand{\ambel}{x}
\newcommand{\configspace}{Q}
\newcommand{\smani}{S}
\newcommand{\mmapping}{\gamma}
\newcommand{\mapvector}{\Gamma}
\newcommand{\shood}{U}
\newcommand{\cshood}{V}
\newcommand{\scoordchart}{\varphi}
\newcommand{\cscoordchart}{\psi}
\newcommand{\csel}{\bm{q}}
\newcommand{\sel}{s}
\newcommand{\kamb}{k^{\ambspace}}
\newcommand{\kconfig}{k^{\configspace}}
\newcommand{\ks}{k^{\smani}}
\newcommand{\kcenters}{\xi}
\newcommand{\coefs}{\hat{{a}}}
\newcommand{\coefsvector}{\hat{\bm{A}}}
\newcommand{\sampPair}{\bm{z}}
\newcommand{\sampSpace}{\mathbb{Z}}
\newcommand{\rkhspace}{\mathbb{H}}
\newcommand{\rkhelement}{h}
\newcommand{\sampvector}{\bm{\configspace}_{\rho}}
\newcommand{\sampelement}{\csel_{\rho}}
\newcommand{\numsamp}{m}
\newcommand{\numfunc}{n}
\newcommand{\numgait}{p}
\newcommand{\dimension}{d}
\newcommand{\ksmatrix}{\bm{K}^\smani}
\newcommand{\projecto}{\Pi}
\newcommand{\lbasis}{\chi}
\newcommand{\RR}{\mathbb{R}}
\newcommand{\ZZ}{\mathbb{Z}}
\newcommand{\HH}{\mathbb{H}}
\newcommand{\spartition}{\{{\smani_i}\}_{i=1}^\numfunc}
\newcommand{\scollection}{\{\sel_i\}_{i=1}^\numsamp}
\newcommand{\cscollection}{\{\csel_i\}_{i=1}^\numsamp}
\newcommand{\EE}{\mathbb{E}}

\newtheorem{thm}{Theorem}

\begin{document}

\title{Learning Theory for Estimation  of \\ Animal Motion Submanifolds}


\author{Nathan Powell and Andrew J. Kurdila
\thanks{Nathan Powell and Andrew J. Kurdila are with the Department of Mechanical Engineering, Virginia Tech, Blacksburg VA 24060, USA
        {\tt\small nrpowell@vt.edu, kurdila@vt.edu}}}

\maketitle
\begin{abstract}
This paper describes the formulation and experimental testing of a  method to estimate submanifold models of animal motion. It is assumed  that the animal motion is supported on a configuration manifold, $Q$, and that the manifold is homeomorphic to a known smooth, Riemannian manifold, $S$. Estimation of the configuration submanifold is achieved by approximating  an unknown mapping, $\gamma$, from $S$  to $Q$.  The overall problem is cast as a distribution-free learning problem over the manifold of measurements. This paper introduces linear approximation spaces in $L^2_\mu(S)$ that contain approximations of $\gamma$ corresponding to those known for classical distribution-free  learning theory over Euclidean space. This paper concludes with a study and discussion of the performance of the proposed method using samples from recent reptile motion studies.
\end{abstract}

\setlength{\intextsep}{0.5pt}
\setlength{\textfloatsep}{0.5pt}
\setlength{\floatsep}{0.5pt}

\section{Introduction}

The related problems of understanding the underlying dynamics of animal locomotion, constructing (multibody) dynamics models of animals, or building robots that emulate animals have occupied researchers having diverse backgrounds. Research in \cite{Bender2015,Birn-Jeffery2016,hudson2012high,Ingersoll2018,porro2017inverse} includes 
investigations related to,  or based upon,  the  motions of humans, bats, birds, lizards, geckos, sheep, frogs, beetles, and cheetahs. Some studies such as in \cite{Brossette2018,Cesic2016,hatton2015nonconservativity,Shammas2007}  seek to understand and utilize the underlying geometry of the mechanics of human or other animal motion. Others are expressly interested in the construction of low dimensional models of complex motion  \cite{bender2017gaussian,rieser2019geometric,wang2007gaussian}. Some are dedicated to finding how qualitative understandings of animal motion can be used to build robots that emulate some observed motion \cite{bayandor2013adaptive,horvat2017spine,karakasiliotis2016cineradiography, shi2018modified,VanTruong2012}, with a notable emphasis on locomotion in quadrupeds \cite{li2011research,seok2013design,semini2016design, Raibert2008,  ugurlu2013dynamic} and bipeds \cite{collins2005bipedal,koolen2016design,negrello2016walk,ott2011posture,radford2015valkyrie,stephens2010dynamic}.

A fundamental question that is either motivated or directly addressed by these studies is how to come up with a general method for defining  a principled, low-dimensional model of a complex animal motion. None of the aforementioned references describe or derive a general strategy that is applicable, in principle, across a broad range of animal motion studies. Just as importantly, none of the references establish rates of convergence that would give confidence in estimates of such low dimensional systems. The aim of our research is to provide further insight into the motion of animals by deriving such a  general method to discover and approximate an  underlying manifold on which the dynamics evolves.

The theory that has been used to derive models of animal motion has been around for some time. These include general multibody dynamics formulations like \cite{shabana2020dynamics}, or the more recent geometric methods \cite{lynch2017modern,bulloLewis2004geometric}. 
As discussed carefully in \cite{bulloLewis2004geometric}, one of the primary theoretical advantages of  geometric methods is their abstraction: states of the system are known to be elements of manifolds, which are by their nature coordinate-free. This generality is particularly attractive in the problem at hand. Intuitively,   we want to approximate or identify some ``lower-dimensional  mathematical object'' that underlies an observed motion of a complex, higher order multibody system. We want to ensure that any associated algorithm converges to a coordinate independent entity.   
In this paper  we introduce the  analysis and experimental testing of a method to identify and approximate motions over a submanifold $Q$ that underlies a particular motion regime observed during animal  motion. We work to formulate the problem in such a way that we can guarantee that approximations converge, in some appropriate sense,  over  the underlying manifold. 

We begin in Section 
\ref{sec:learning_manifolds} with a careful description of the problem of estimating a motion manifold as one of distribution-free learning over manifolds. This problem is substantially more complicated than the conventional problems for functions defined  over Euclidean spaces. This new learning problem is made complex in no small part owing to the difficulty of defining appropriate function spaces and spaces of approximants over manifolds. These two topics are covered in Sections \ref{sec:functions_manifold} and \ref{sec:approximant_manifold}, respectively. In particular, we introduce the linear approximation spaces $A^{r,2}(L^2_\mu(S))$ for a smooth manifold $S$, and these play a crucial role in deriving rates of convergence. The primary theoretical result of this paper gives sufficient conditions to ensure that approximations $\gamma_{n,m}:S\rightarrow X$ of the unknown function $\gamma:S\rightarrow Q\subset X$ converge to the regressor function $\gamma_\mu:S\rightarrow X$. Additionally, we show that when the regressor function satisfies the smoothness condition $\gamma_\mu\in (A^{r,2}(L^2_\mu(S)))^d$, the rate of convergence is given by 
{\vspace*{-1em}
\begin{align}
   \EE( \|\gamma^j_\mu-\gamma^j_{n,m}\|_{L^2_\mu}) 
    &\leq C_1 (N(n))^{-r} + C_2 \frac{N(n)\log(N(n))}{m}
    \label{eq:th_bound}
\end{align}}
\noindent for constants $C_1,C_2>0$, where $\EE(\cdot)$ is the expectation over samples, $N(n)$ is the dimension of the space of approximants, and $m$ is the number of samples. This is precisely the rate achieved for certain approximations of distribution-free learning problems over Euclidean space. Such rates of convergence are known to be optimal in the Euclidean case, except for the logarithmic factor. \cite{devore2006approximation}  Finally, this paper concludes in Section \ref{sec:experiment}  with a study of the performance of the proposed method using samples from recent reptile motion studies.

\section{Problem Formulation} 

\subsection{Kinematics and Inherent Geometry}
\label{sec:kinematics}


For purposes of ensuring convergence of approximations, we assume in this paper that motions are supported on a configuration manifold, $\configspace$, that is contained in the ambient space  $X\approx \RR^d$  for some $d>0$.   Of course this space suffices to describe the dynamics of multibody systems comprised of assemblies of lumped masses where motion is specified in terms of the mass centers of the  (inertia-free) bodies. Since any Riemannian manifold can be (isometrically) embedded in $\RR^d$ for a large enough $d$ via the Nash embedding theorem, in  principle this assumption also allows for some of the other standard models where the configuration manifold includes SO(3) or SE(3) as described in \cite{bulloLewis2004geometric,lynch2017modern}.  
 It is further assumed that the manifold $\configspace$ is a smooth, compact, connected, Riemannian manifold that is regularly embedded in the ambient space $\ambspace \approx \mathbb{R}^d$ This manifold is taken to be homeomorphic to some known smooth manifold $\smani$.  We denote the  homeomorphism  by  $\mmapping: \smani \to \configspace$ where
$\gamma:=[\gamma^1,\ldots,\gamma^d]^T$ and $\gamma^i:S\rightarrow \RR$, and we discuss its smoothness in our discussion of approximation spaces in  Section \ref{sec:functions_manifold}. 
The manifold  $\smani$ is equipped with the topology induced by its intrinsic Riemannian metric and $\configspace$ is equipped with the topology inherited as a regularly embedded submanifold  $Q \subseteq X\approx  \mathbb{R}^d$.

\subsection{The Distribution-Free Learning Problem on Manifolds} 
\label{sec:learning_manifolds}
This paper is concerned with finding or approximating the mapping $\gamma$ that determines the low-dimensional underlying manifold $Q$ that supports a given motion or motion regime. 
We choose to formulate this problem as one of {\em distribution-free} learning theory on manifolds. In distribution-free learning theory it is assumed that we are given a collection of independent and identically distributed samples $\{z_i\}_{i=1}^m:=\{(s_i,q_i)\}_{i=1}^m\subset \ZZ:=S \times X$ that that are generated by some {\em unknown} probability measure $\mu$ on the manifold $\ZZ$. 
In order to determine the mapping from the known  manifold $\smani$ to the configuration space $\configspace$,  it would be ideal if we could determine an optimal mapping $\hat{\gamma}:S\rightarrow Q\subset X$ such that 
\vspace*{-.75em}
\begin{equation}
    \hat{\mmapping} =
    \argmin_{\gamma\in \Gamma} E_\mu(\gamma):= \argmin_{\mmapping\in \Gamma}  \int_\ZZ \|q - \mmapping(\sel)\|_{X}^2\mu(d\sampPair)
    \label{eqn:distFreeLearning}
\end{equation}
where $z=(s,q)\in \ZZ$ and $\Gamma$ is a set of admissible operators that map from $S\rightarrow Q$. 
 Solving the optimization problem \ref{eqn:distFreeLearning} poses a number of difficulties, some of which are rather standard challenges in the field of distribution-free learning theory. For example, it is well-known that the ideal solution $\hat{\gamma}$ above cannot be computed from the minimization problem in Equation \ref{eqn:distFreeLearning} since the measure $\mu$ on $\ZZ$ is unknown. For this reason methods of empirical risk minimization are used that substitute a discrete proxy for the expression in Equation 
 \ref{eqn:distFreeLearning}. We discuss this step in detail shortly in Section  \ref{sec:empirical_risk}. 
 
 Beyond these conventional challenges to building approximations of $\hat{\gamma}$, there are additional substantial difficulties that are unique to the problem at hand. In classical treatments of distribution-free learning theory, the rates of convergence of approximations to the solution are often cast in terms of approximation spaces or smoothness spaces (that often end up being  Sobolev spaces)  of real-valued functions over subsets of Euclidean space  $\RR^p$. That is, the space $\Gamma$ is usually selected to be some subset of the real-valued functions over subsets of Euclidean space. Here, however, the functions $\gamma\in \Gamma$ are mappings from the manifold $S$ to the manifold $Q\subset X$. The definition of Sobolev mappings between manifolds is a complicated subject, and for general choices of $S$ or $Q$ it is unclear exactly what definition of smoothness should be selected. A significant part of this paper is dedicated to defining  approximation spaces, $A^{r,2}(L^2_\mu(\smani))$, over manifolds to characterize smoothness or approximation properties and subsequently   structuring the learning problem so convergence  results can be derived in a way that is analogous to the classical case over subsets of Euclidean Space. 

\subsection{Function Spaces on  Manifolds}
\label{sec:functions_manifold}

In the setup of the problem above we have assumed that $S$ is a known, compact, connected, smooth Riemannian manifold, and that $Q$ is a regularly embedded submanifold of $X:=\RR^d$. This means that the unknown function $\gamma:S\rightarrow Q$ is constructed of   component functions $\gamma^i:S\rightarrow \mathbb{R}$ for $i=1,\ldots,d$. In this section we define the various function spaces over manifolds that will be used to approximate the functions $\gamma^i$. 
The function spaces over the manifold $S$  that is used in this paper consist of certain spaces of square-integrable functions and native spaces of a reproducing kernel. When $\mu_S$ is a measure over the manifold $S$, the usual space of real-valued, $\mu_S$-square integrable functions is given by $L^2_\mu:=L^2_{\mu_S}(S)$, which is a Hilbert space with respect to the usual inner product $(f,g)_{L^2_{\mu}}:=\int_S f(\xi)g(\xi)\mu_S(d\xi)$.  Recall \cite{saitoh2013reproducing,berlinet2011reproducing} that a real RKH space $\HH^A$ over an arbitrary set $A$ is defined to be 
$
    \HH^A:=\overline{\text{span} \{
    k^A_a \ | \ a \in A \} } 
$ 
where $k^A_a:=k^A(a,\cdot)$ and $k^A:A\times A \rightarrow \RR$ is a real, positive semi-definite,  symmetric, continuous admissible kernel function. This definition of an RKH space makes sense over a general set $A$, and in particular makes sense for the specification of functions over manifolds.  
We  denote by  $\kamb: \ambspace \times \ambspace \to \mathbb{R}$ the kernel of  the RKH Space $\rkhspace^{\ambspace}$ over the set  $\ambspace$. The restriction of functions in $\rkhspace^\ambspace$ to the set $Q$ always defines an RKH space over $Q$, and we set 
$
   \rkhspace^{\configspace} :=  \left \{ \; g \rvert_{\configspace} \; \    | \  g  \in \rkhspace^{\ambspace} \right \}$.
The kernel $\kconfig: \configspace \times \configspace \to \mathbb{R}$ that defines the space of restrictions $\HH^Q$  is given by
$   \kconfig(x,y) := \kamb(x,y) \rvert_{\configspace}$, for all $ x,y \in \configspace$, 
which is a classical result from the theory of RKH spaces. 
Finally, we define the pullback space
$\HH^S:=\mmapping(\rkhspace^{\configspace})$ on $\smani$ as
$ 
   \mmapping^{\smani}(\rkhspace^{\configspace}) := \{ \; g \circ \mmapping \; | g  \in \rkhspace^{\configspace} \},
$ 
and  the kernel $k^S$  defined on the manifold $\smani$ is written as 
$
   \ks(\alpha,\beta) = \kconfig(\mmapping(\alpha),\mmapping(\beta)).
$

\subsection{Approximant Spaces} 
\label{sec:approximant_manifold}
The approximations in this paper are constructed in terms of two different types of finite dimensional spaces of approximants. 
\subsubsection{Approximants in $L^2_\mu(S)$}
When we build approximations in $L^2_\mu(S)$, we make use of a nested sequence $\{\mathcal{S}_n\}_{n=1}^\infty$ of measurable partitions  $\mathcal{S}_n$ of $S$, where each partition  $\mathcal{S}_n:=\{{S}_{n,k} \}_{k=1}^{N(n)}$.
 That is, these  sets   satisfy  $S=\bigcup_{k=1}^{N(n)} {S}_{n,k}$,  ${S}_{n,k}\bigcap {S}_{n,\ell}=\emptyset$ for all $\ell\not= k$, and $S_{n,k}=\bigcup_{\ell\in \Lambda_{n,k}} S_{n+1,\ell}$ for some finite set of indices $\Lambda_{n,k}$. 
 Here $N(n)=\#(\mathcal{S}_n)$ is the number of sets in the $n^{th}$ partition $\mathcal{S}_n$. 
We define the space of approximants $H^S_n$ as the span of the characteristic functions $1_{S_{n,k}}$ of the sets in $\mathcal{S}_n$, 
$
    H^S_n := \text{span}\{1_{\smani_{n,k}}  \  | \ 1 \leq k \leq N(n) \} \subset L^2_\mu(S) 
    \label{eqn:spanset}
$
where $\{1_{S_{n,k}}\}_{k=1}^{N(n)}$ is the associated partition of unity over $S$. 
We associate to the partition $\mathcal{S}_n$ a set of representatives, 
$
\Xi_n:=\left 
\{\xi_{n,k} \ | \ \xi_k\in S_{n,k}, k=1,\ldots, N(n)\right \}.
$ 
These representative points  are assumed to fill the manifold in the sense that  the fill distance 
$
    h_{\Xi_n,S}:=\max_{s\in S} \min_{\xi_{n,k}\in \Xi_n} d(s,\xi_{n,k}) \rightarrow 0
$ 
as $n\rightarrow \infty$. 
For any $g \in L^2_\mu(s)$ we define the $L^2_\mu$-orthogonal projection $\Pi^S_n$ in the usual way, 
\begin{align*}
     \projecto_\numfunc^\smani g(\cdot) &= \sum_{k=1}^{N(n)}\int_{S_{n,k}} g(\ambel)\frac{1_{\smani_k}(\sel) \mu_\smani(d\sel)}{\sqrt{\mu_\smani (\smani_{n,k})}}\frac{1_{\smani_{n,k}}(\cdot)}{\sqrt{\mu_\smani (\smani_{n,k})}}. 
\end{align*} 
{\noindent We define the linear approximation space 
$A^{r,2}(L^2_\mu)\subseteq L^2_\mu(S)$ of functions defined over the manifold $\smani$ as
{
\small
\begin{equation*}
A^{r,2}(L^2_\mu)=\left \{ g\in L^2_\mu(S) \ \biggl  | \ \exists C \text{ s.t. }  \|(I-\Pi^S_n)g\|_{L^2_\mu} \leq C(N(n))^{-r}
\right \}.
\end{equation*}
} 
\noindent The infimum of the constants $C$, for which in the inequality above holds, defines a semi-norm on $A^{r,2}(L^2_\mu)$. We note that if $\smani$ is a compact subset of Euclidean space and $\mu$ is a Lebesgue measure, these spaces can be understood in some cases as Sobolev spaces \cite{devore2006approximation}. Here the expression above defines approximation spaces over the manifold $\smani$.}

\subsubsection{Approximants in $\HH^S$}
Approximations are also constructed in terms the finite dimensional spaces of approximants
$
\HH_m^S:= \text{span}\{ k^S_{s_i} \ | \ 1\leq i \leq m  \} \subset \HH^S
$ and $
\HH_m^Q:= \text{span}\{ k^Q_{s_i} \ | \ 1\leq i \leq m \}\subset \HH^Q,
$ 
that are defined in terms of the samples $\{(s_i,x_i)\}_{i=1}^m\subset \ZZ$. The solutions generated from this space are given by the data-driven EDMD method discussed in \cite{klus2020} and \cite{brunton2019data}.
%
%
\subsection{Empirical Risk Minimization over Manifolds}
\label{sec:empirical_risk} 
As noted above, in some ideal sense the optimal choice of the mapping $\gamma:S\rightarrow Q$ is the one that extremizes the cost functional $E_\mu$ in Equation \ref{eqn:distFreeLearning}. 
Since the measure $\mu$ that generates the samples in $\ZZ$ is unknown, distribution-free learning theory uses a proxy that can be computed to re-cast the optimization problem.  The form of the empirical risk can best be motivated by first  rewriting the ideal risk $E_\mu$ in terms of the regressor function. 
Any measure $\mu$ on the space of samples $\ZZ$ can be rewritten (disintegrated) into the expression 
$\mu(dz):=\mu(dxds):=\mu(dx|s)\mu_S(ds)$  with $z:=(s,x)$,  $\mu_S$ the associated marginal measure over $S$, and $\mu(\cdot | s)$ the conditional probability measure on $X$ given $s\in S$.  From first principles it is straightforward to show that we can rewrite the ideal cost  $E_\mu$ in the alternative form 
\begin{equation}
\label{eq:dfl_2}
E_\mu(\gamma)=\int_S \|\gamma(s)-{\gamma}_\mu(s)\|^2 \mu_S(ds) + E_\mu({\gamma_\mu})
\end{equation}
where 
$ 
\gamma_\mu(s):=\int_X x \mu(dx | s)
$
is the regressor function associated with the measure $\mu(dz):=\mu(dx|s)\mu_X(ds)$. This decomposition shows, in fact, that the optimal mapping $\hat{\gamma}$ that minimizes $E_\mu(\mmapping)$ is the regressor function $\hat{\gamma}:=\gamma_\mu$. 
Since the last term on the right  in Equation \ref{eq:dfl_2} is a constant that does not depend on $\gamma$, we introduce the empirical risk 
\begin{equation}
    E_m(\gamma) = \frac{1}{m}\sum_{i=1}^m
    \| \ambel_i - \mmapping(\sel_i)\|_\ambspace^2
    \label{eqn:empRiskMin}
\end{equation}
in terms of the samples $\{(s_i,x_i)\}_{i=1}^m\subset \ZZ$. Note that the cost functional $E_m(\gamma)$ can be computed for any admissible function $\gamma$ since the samples are known. We will be interested in two types of approximations of the empirical risk in the discussions that follow. We define 
\begin{equation}
   \gamma_{\numfunc,\numsamp}= \argmin_{\gamma_n\in H^S_n}\frac{1}{m}\sum_{i=1}^\numsamp ||\ambel_i - \mmapping_\numfunc(\sel_i)||_\ambspace^2
    \label{eqn:minimizer1}
\end{equation}
when we seek approximations that converge in $L^2_\mu$. This definition has the advantage that it is possible to derive strong convergence rates for the approximations $\gamma_{n,m}\rightarrow \gamma_\mu$. However, the resulting approximations are not very smooth.  Alternatively, we replace $H_n^S$ by $\HH^S_n$ in Equation \ref{eqn:minimizer1}  
when we seek smoother representations of the mapping $\gamma_{n,m}$. As of yet, we have not derived such strong rates of convergence $\gamma_{n,m}\rightarrow \gamma_\mu$ in this case. However, numerical examples compare the performance of the estimates obtained with the choices $H^S_n$ and $\HH^S_n$. 

Before we study the solutions of these distribution-free learning problems over the manifold $S$, we make one last observation regarding the form of minimization problem. 
By introducing the samples 
$x_i:=\{x_i^1,\ldots,x_i^d\}^T$ and mapping $\gamma_n:=\{\gamma_n^1,\ldots,\gamma_n^d\}^T$, we can write 
\begin{align*}
    \gamma_{\numfunc,\numsamp}&= \argmin_{\gamma_n\in H^S_n}\frac{1}{m}\sum_{i=1}^\numsamp \sum_{j=1}^\dimension (\ambel^j_i - \mmapping^j_\numfunc(\sel_i))^2 = \argmin_{\gamma_n\in H^S_n}\sum_{j=1}^\dimension E_\numsamp^j(\mmapping_\numfunc^j)
\end{align*}
with $E^j_m(\gamma_n^j):=\frac{1}{m}\sum_{i=1,\ldots,m}(x_i^j-\gamma_n^j(s_i))^2$. 
Note that since each term $E^j_m(\gamma_n^j)$ is positive, the optimal $\gamma_{n,m}$ can be obtained by extremizing each of the terms $E^j_m(\gamma_n^j)$ for the component functions $\gamma_n^j$ for $j=1,\ldots, d$. The primary theoretical result of this paper is summarized in the following theorem, which solves the distribution free learning problem over manifold $\smani$
\begin{thm}
\label{th:error_rate}
The optimal solution $\gamma_{n,m}:=[\gamma^1_{n,m},\ldots,\gamma^{d}_{n,m}]^T$ of the 
empirical risk minimization problem in Equation \ref{eqn:empRiskMin} is given by 
\begin{equation*}
    \mmapping_{\numfunc,m}^{j}(s) = \sum_{k=1}^n\frac{\sum_{i=1}^m1_{\smani_k}(\sel)\ambel_i^{j}}{\sum_{i=1}^m1_{\smani_k}(\sel)} 1_{\smani_k}(\sel).
\end{equation*} 
If the regressor $\gamma_\mu\in A^{r,2}(L^2_\mu(S))$, we have the error bound
\begin{align}
   \EE( \|\gamma^j_\mu-\gamma^j_{n,m}\|_{L^2_\mu}) 
    &\leq C_1 (N(n))^{-r} + C_2 \frac{N(n)\log(N(n))}{m}
    \label{eq:th_bound2}
\end{align}
where the constant $C_1$ is independent of $\gamma$ and $C_2=C_2(\gamma)$ and $\EE(\cdot)$ is the expectation on $S^m$ with the product measure $\mu_S^m$.  
\end{thm} 
\begin{proof}
When we define $\gamma^j_{n}:=\sum_{k=1,\ldots,n}\hat{a}^j_{n,k} 1_{S_k}(\cdot)$, we have the explicit representation of $E^j_m(\gamma^j_{n})$ given by 
\begin{equation*}
    E_\numsamp^j(\mmapping_\numfunc^j) = \frac{1}{m}\sum_{i=1}^m\left (\ambel_i^j - \sum_{k=1}^n\coefs_{\numfunc,k}^j 1_{\smani_k}(\sel_i)\right )^2.
\end{equation*}
 We can also write this sum as
\begin{equation*}
    E_\numsamp^j(\mmapping_\numfunc^j) = \frac{1}{m}\sum_{k=1}^n\sum_{i=1}^m\left (1_{\smani_k}(\sel_i)\ambel_i^j - \coefs_{\numfunc,k}^j 1_{\smani_k}(\sel_i)\right )^2,
\end{equation*}
and this summation can be reordered as 
$ 
    E_\numsamp^j(\mmapping_\numfunc^j) = \frac{1}{m}\sum_{k=1}^nE_{\numsamp,\numfunc}^j(\coefs_{n,k}^j)
$ and 
\begin{equation*}
   E_{\numsamp,\numfunc}^j(\coefs_{n,k}^j) = \sum_{i=1}^m(1_{\smani_k}(\sel)\ambel_i^j - \coefs_{\numfunc,k}^j1_{\smani_k}(\sel))^2
\end{equation*}
with each $E_{m,n}^j(\hat{a}^j_{n,k})$ depending on a single variable $\hat{a}^j_{n,k}$. 
By taking the partial derivative $\partial(E^j_{m,n}(\hat{a}^j_{n,k}))=0$, we see that 
$ 
    \coefs_{\numfunc,k}^{j} = {\sum_{i=1}^m1_{\smani_k}(\sel_i)\ambel_i^{j}}/{\sum_{i=1}^m1_{\smani_k}(\sel_i)}
    , $
which establishes the form of solution stated above. 
We now turn to the consideration of the error bound in the theorem.  From the triangle inequality
\begin{equation*}
    ||\mmapping_\mu - \mmapping_{\numfunc,\numsamp}||_{L^2_\mu} \leq ||\mmapping_\mu - \projecto_\numfunc^\smani \mmapping_\mu||_{L^2_\mu} + || \projecto_\numfunc^\smani \mmapping_\mu-\mmapping_{\numfunc,\numsamp}||_{L^2_\mu},
\end{equation*} 
we can bound the first term above by $(N(n))^{-r}$ from the definition of the linear approximation space $A^{r,2}(L^2_\mu(S))$. The bound in the theorem is proven if we can show that there is a constant $C_2$ such that $\|\Pi_n^S \gamma_\mu^j - \gamma_{n,m}\|_{L^2_\mu(S)}\leq C_2 N(n)\log(N(n))/m$.
We establish this bound by a straightforward extension to functions on the manifold $S$ of the proof in \cite{binev2005universal}, which is given for functions defined on $\RR^p$ for $p\geq 1$. The expression above for $\gamma^j_{n,m}$ can be written in the form $\gamma^j_{n,m}:=\sum_{k=1}^{N(n)}\alpha^j_{n,k}1_{S_{n,k}}$, and that for $\Pi^S_n\gamma^j_\mu$ can be written as  $\gamma^j_{\mu}:=\sum_{k=1}^{N(n)}\hat{\alpha}^j_{n,k}1_{S_{n,k}}$. In terms of these expansions, we write the error as 
$$
\| \Pi^S_n \gamma^j_\mu-\gamma_{n,m}\|^2_{L^2_\mu(S)}=\sum_{k=1}^{N(n)}
\left ( \alpha_{n,k}^j-\hat{\alpha}_{n,k}^j \right )^2\mu_S(S_{n,k}).
$$
Let $\epsilon>0$ be an arbitrary, but fixed, positive number. We define the  set of indices $\mathcal{I}(\epsilon)$ that denote subsets $S_{n,k}$ that have, in a sense, small measure,
\begin{align*}
    \mathcal{I}(\epsilon)&:= \left\{ k\in \{1,\ldots,N(n)\}\ \biggl | \ \mu_S(S_{n,k})\leq \frac{1}{8N(n)\bar{X}^2} \right \} 
\end{align*}
where $\bar{X}$ is $\sup_{s\in S}\|\gamma(s)\|_X$.  We define the complement  $\tilde{\mathcal{I}}
(\epsilon):=\{ k\in \{1\ldots N(n)\} \ | \ k\not\in \mathcal{I}\}$, and then set the associated sums
$
S_{\mathcal{I}}:=\sum_{k\in \mathcal{I}}(\alpha_{n,k}^j-\hat{\alpha}_{n,k}^j)^2\mu_S(S_{n,k})
$ 
and 
$
S_{\tilde{\mathcal{I}}}:=\sum_{k\in \tilde{\mathcal{I}}}(\alpha_{n,k}^j-\hat{\alpha}_{n,k}^j)^2\mu_S(S_{n,k})
$. 
The bound in Equation \ref{eq:th_bound} follows if we can demonstrate a concentration of measure formula 
\begin{align}
\label{eq:firstline}
&\text{Prob}\left (
 \|\Pi^S_n \gamma^j_\mu
-\gamma^j_{n,m} \|^2_{L^2_\mu} > \epsilon^2 \right )\\
& \hspace{.2in} \notag =\text{Prob}(S_{\mathcal{I}}+S_{\tilde{\mathcal{I}}}>\epsilon^2)\leq 
be^{cm\epsilon^2/N(n)}
\end{align}
for some constants $b,c$. See \cite{binev2005universal,devore2006approximation} for a discussion of such concentration inequalities. The fact that such a concentration inequality implies the bound in expectation in Equation \ref{eq:th_bound} is proven in \cite{binev2005universal} on page 1311 for functions over Euclidean space. The argument proceeds exactly in the same way for the problem at hand by integration of the distribution function defined by Equation \ref{eq:firstline} over the manifold $S$. To establish the concentration inequality, let us define two events 
$
E_{\mathcal{I}+\tilde{\mathcal{I}}}(\epsilon) :=\left \{ z\in \ZZ^m\ | \ S_{\mathcal{I}}+S_{\tilde{\mathcal{I}}}\geq \epsilon^2 \right \}$, 
and $
E_{\tilde{\mathcal{I}}}(\epsilon) :=\left \{ z\in \ZZ^m\ | \ S_{\tilde{\mathcal{I}}}\geq \frac{1}{2}\epsilon^2. \right \}.
$
We can compute directly from the definitions of the coefficients $\alpha^j_{n,k},\hat{\alpha}^j_{n,k}$, and using the the compactness of $\gamma(S)\subset X$, that $S_{\mathcal{I}}\leq \epsilon^2/2$ for any $\epsilon>0$. Since we always have 
{\small 
\begin{align*}
    & \sum_{k\in \tilde{I}}\left (
    \alpha^j_{n,k}-\hat{\alpha}^j_{n,k}
    \right )^2\mu_S(S_{n,k})  >
    \epsilon^2 - \sum_{k\in \mathcal{I}}\left (
    \alpha^j_{n,k}-\hat{\alpha}^j_{n,k}
    \right )^2 \mu_S(S_{n,k})
\end{align*}
}
\noindent is bounded below by $> \frac{1}{2}\epsilon^2$, 
we know that $E_{\mathcal{I}+\tilde{\mathcal{I}}}(\epsilon)\subseteq E_{\tilde{\mathcal{I}}}(\epsilon)$ for any $\epsilon>0$. If the inequality $S_{\tilde{\mathcal{I}}}>\epsilon^2/2$, then we know there is at least one $\tilde{k}\in \tilde{\mathcal{I}}$ such that 
\begin{align*}
    S_{\tilde{k}}(\epsilon)
    :=(\alpha^j_{n,k}-\hat{\alpha}^j_{n,k})^2\mu_S(S_{n,k})>\frac{1}{2 (\#\tilde{\mathcal{I}})}\epsilon^2>\frac{1}{2N(\epsilon)}\epsilon^2. 
\end{align*}
When we define the event 
$ E_k(\epsilon):=\{z\in \ZZ^m \ | \ S_k(\epsilon)> \epsilon^2/2N(n) $ for each $k\in \{1,\ldots, N(n)\}$, we conclude 
$
E_{\mathcal{I}+\tilde{\mathcal{I}}}\subseteq E_{\tilde{\mathcal{I}}}(\epsilon) \subseteq \bigcup_{k\in \tilde{\mathcal{I}}} E_k. 
$
By the monotonicity of measures, we conclude that 
$\text{Prob}\left (\mathcal{I}+\tilde{\mathcal{I}} \right ) \leq \sum_{k\in \tilde{\mathcal{I}}} \text{Prob}(E_k)$. But we can show, again by a simple modification of the arguments on page of \cite{binev2005universal}, that $\text{Prob}(E_k)\lesssim e^{-cm\epsilon^2/N(m)}$. The analysis proceeds as in that reference by using Bernstein's inequality for random variables defined over the probability space $(S,\Sigma_S, \mu_S)$ instead of over Euclidean space.

\end{proof}


\section{Experiments and Numerical Examples}
\label{sec:experiment}
\begin{figure}[h!]
\centering
    \includegraphics[width =0.22\textwidth]{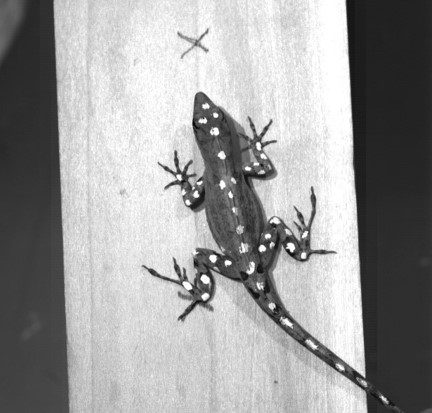}
\caption{Specimen {\em anolis sagrei} with markers.}
\label{fig:LizardMarkers}
\end{figure}

A number of experiments and simulations have been carried out to study the performance of the analysis in the paper. Photoreflective markers were placed on specimens of {\em anolis sagrei} as depicted in Figure \ref{fig:LizardMarkers}, which then traversed an inclined, narrow board within the line of sight of three high speed, Photron FASTCAM \textregistered cameras. Video imagery was recorded at 500 fps, and inertial trajectories such as those depicted in Figure 2 of the markers were estimated using the publicly distributed software toolbox \cite{hedrick2008software}. 
 The discrete inertial trajectories are interpreted as measured samples $\{(t_i,x_i)\}_{i=1}^m \subset [0,\infty)\times X$. In this example, we hypthesize  that the observed configuration space is homeomorphic to a  one-dimensional, compact space. Consequently, we choose the known manifold $S=S^1$, the circle in $\RR^2$, and we seek a mapping $\gamma:S\rightarrow Q\subset X$ where $Q$ is the unknown manifold homeomorphic to $S$. 
 The approximation   $\gamma_{n,m}$ is obtained as the solution of  the empirical risk minimization problem  for two choices $H^S_n$ and $\HH^S_n$ of approximant subspaces in the plots that follow. The basis for $\HH^S_n$ is taken to be given by translates of the exponential kernel  $k_{\xi}(s) = e^{-\beta||\xi-s||^2} $. Figures 3 and 4 compare the performance of the two methods of approximation as the number of samples and the dimension of the approximation spaces are varied.

Figure \ref{fig:gridTesting} illustrates the effects that the number of centers has on the fidelity of  the approximation. For both methods, it is clear that increasing the dimension number $N(n)$ of partitions corresponds to a higher resolution that manages to capture changes in the data over smaller intervals. 
The EDMD approximations generated from the exponential kernel  are  also dependent on the hyperparameter $\beta$. 

\begin{figure}[h!t]
\centering
    \includegraphics[width =0.4\textwidth]{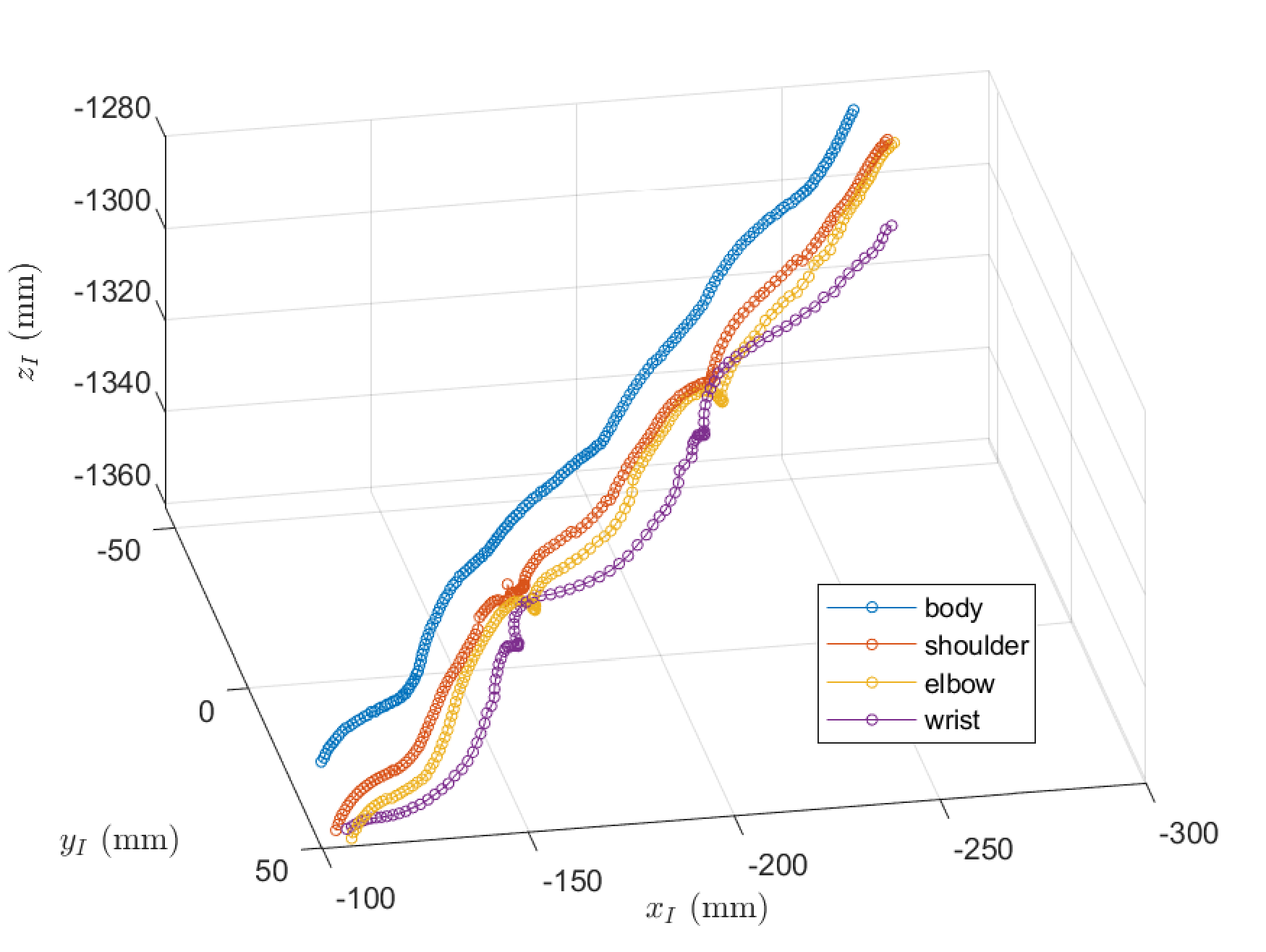} 
\caption{Inertial Trajectories of the markers placed on a lizard in three-dimensional space.}
\label{fig:InertialTrajectories3D}
\end{figure}
Figure \ref{fig:betaTesting} illustrates  the effects of the hyperparameter $\beta$ on the approximation. For an exponential basis, larger values of $\beta$ result in basis functions that decay rapidly from their centers,  "sharpening" the curve. This results in a more oscillatory approximation from the basis functions. Thus,while the optimization of the empirical risk for functions in $H^S_n$ yields nonsmooth estimates, there is no analog to the hyperparameter $\beta$. Depending on the hyperparameter $\beta$, highly oscillatory solutions can result for different choices of $\beta$, even for EDMD solutions over approximant spaces having the same dimension $N(n)$. Such oscillations can be addressed in the EDMD methods or Gaussian process models by introducing a regularization parameter. \cite{klus2020}

\begin{figure}[h!t]
\centering
    \includegraphics[width =0.4\textwidth]{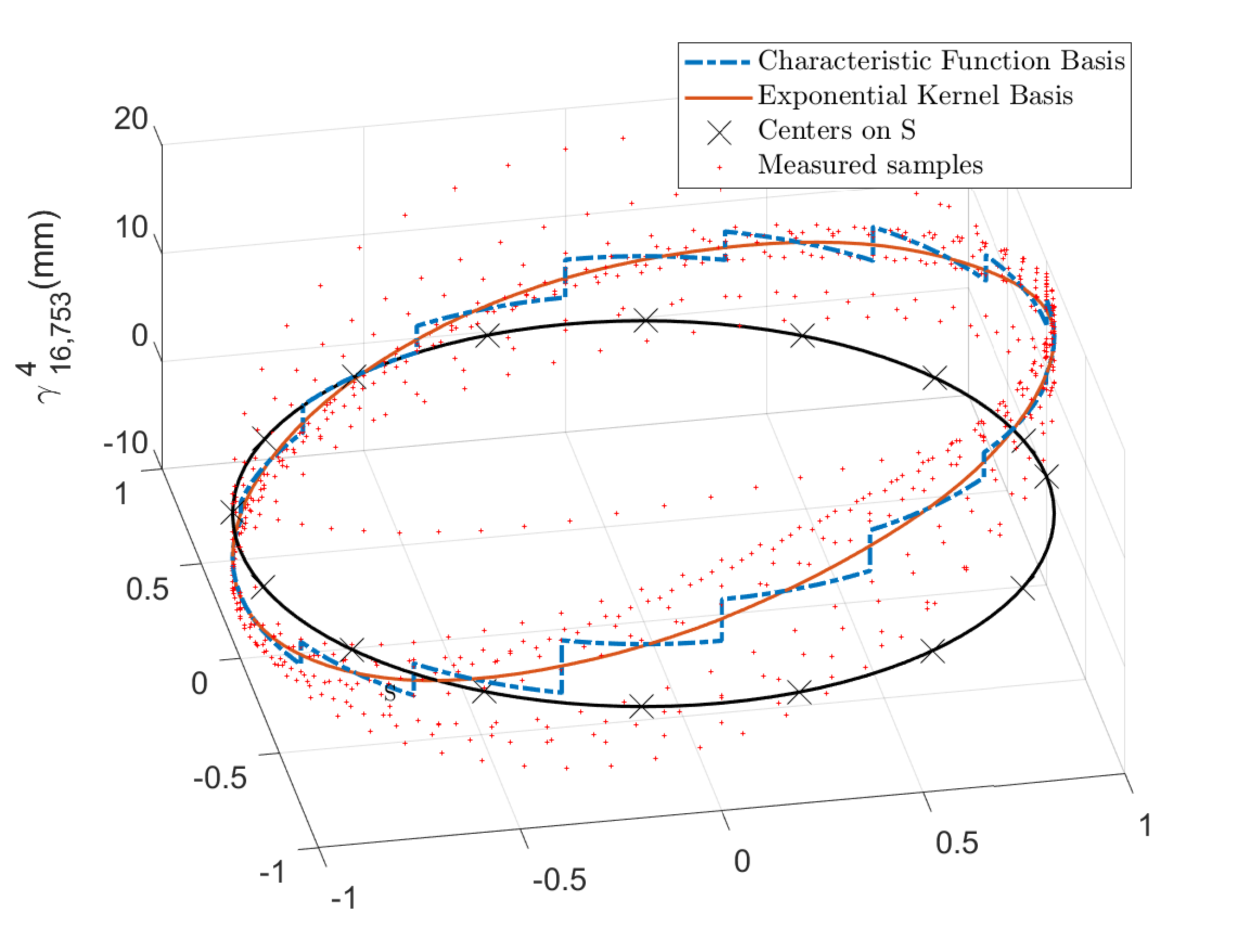}
\caption{Comparison of  two methods of approximation of the unknown map $\gamma:S\rightarrow Q$ for a fixed number of samples, $m=32$. 
}
\label{fig:methodTesting}
\end{figure}

\begin{figure}[h!t]
{\includegraphics[width =0.45\textwidth]{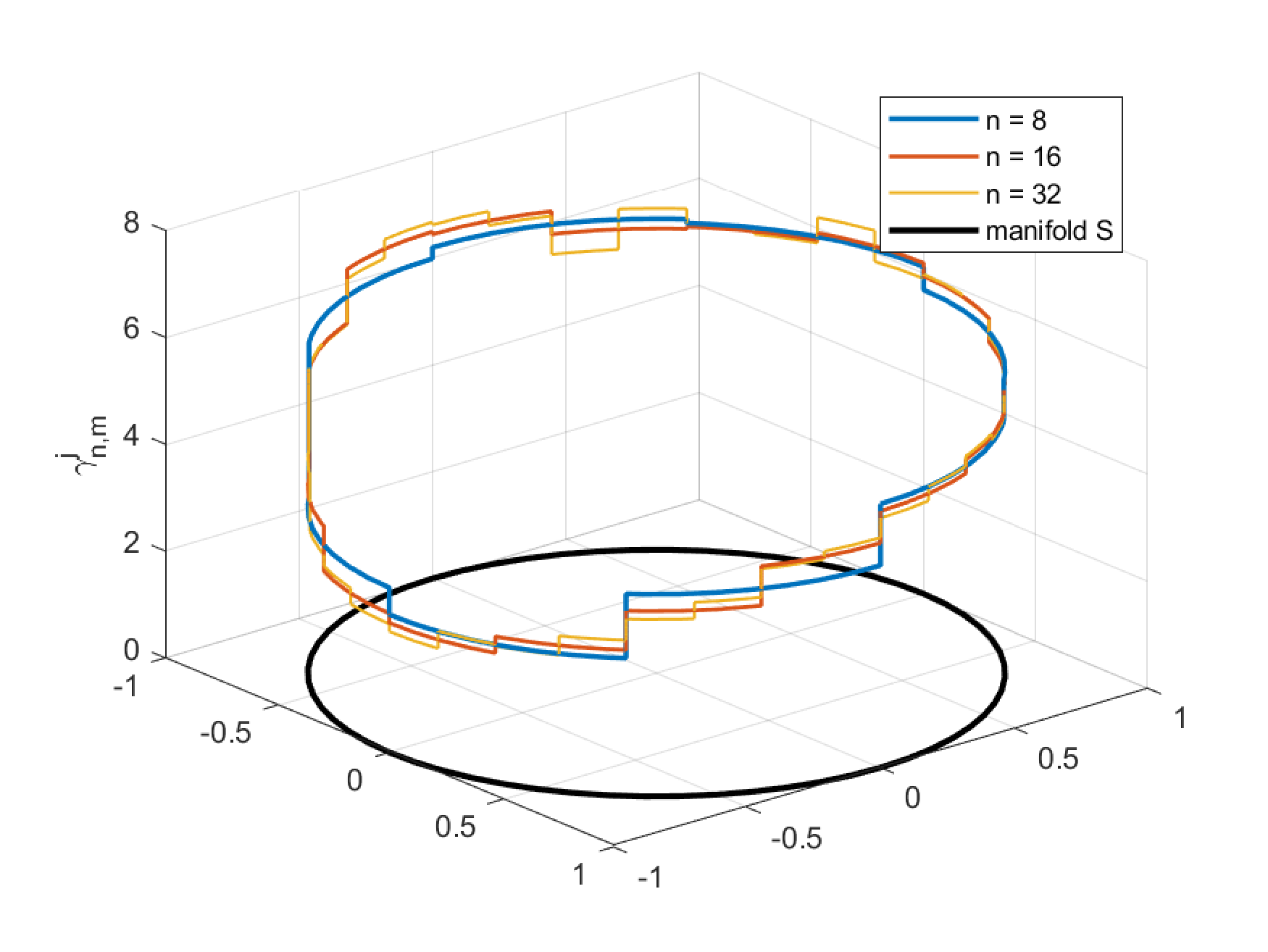}}
{\includegraphics[width =0.45\textwidth]{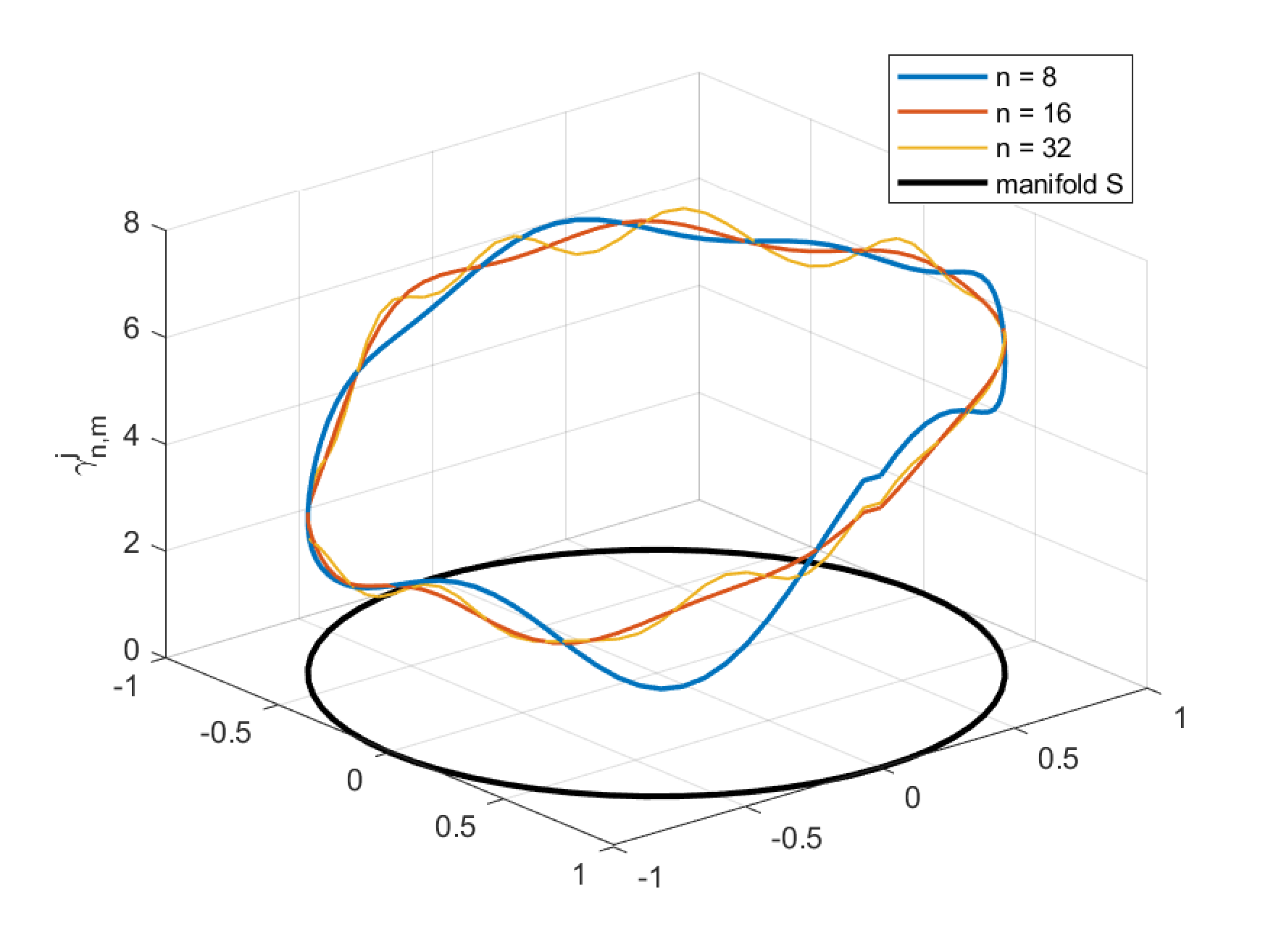}}
\caption{The approximations of the mapping $\gamma^j$ from (top) the partition $\mathcal{S}_n$ using the characteristic basis  $1_{\smani_{n,k}}$,  and (bottom) the exponential kernel basis in $\HH^S_n$. 
}
\label{fig:gridTesting}
\end{figure}

\begin{figure}[ht]
\centering
    \includegraphics[width =0.45\textwidth]{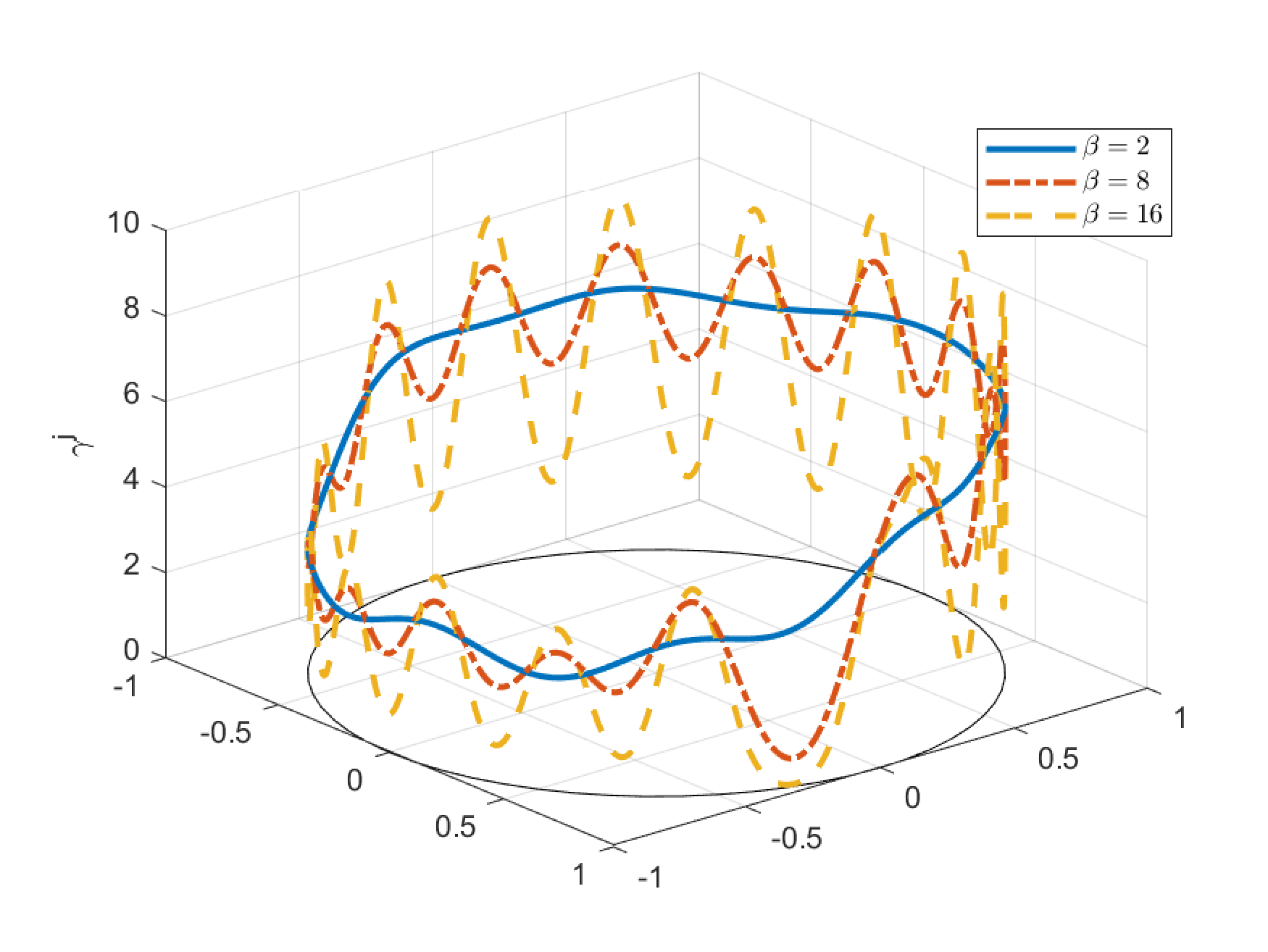}
\caption{The approximations using a fixed number of $N(n)=16$ centers for various values of the kernel hyperparameter, $\beta$.
}
\label{fig:betaTesting}
\end{figure}

\section{Conclusions}
This study has derived  a general formulation of the problem to approximate  a submanifold representing animal motion from observations. The general form of the problem has been stated as a type of distribution-free learning over a smooth manifold. The paper defines  a closed form expression for approximations 
$\gamma_{n,m}$ of the map $\gamma:S\rightarrow Q$ that defines the manifold, and derives bounds for the rates of convergence over the manifold that are analogous to those obtained in corresponding problems over Euclidean spaces.  

\section*{Appendix}
Let the sample pairs be deonted $\{(\sel_i,\csel_i)\}_{i=1}^m \subseteq \smani \times \configspace \subseteq \smani \times \ambspace$. Where $\mmapping(\sel_i) = \csel_i$. By definition 
\begin{equation*}
      \kconfig \sim \rkhspace^{\configspace} \qquad
\end{equation*}
\begin{equation*}
   \bigg{\{}\begin{array}{cccccc}
         \configspace_\numsamp = & \{\csel_i &| 1 \leq i \leq \numsamp\}  \\
         \rkhspace^{\configspace}_\numsamp = & \text{span} \{\kconfig_{\csel_i} &| \csel_i \in \configspace_\numsamp\} =& \text{span}\{\lbasis_{\csel_i}^\configspace &| \csel_i \in \configspace_\numsamp \} \\
         \projecto_\numsamp^{\configspace} :& \rkhspace^{\configspace}  \to \rkhspace^{\configspace}_\numsamp&\  &
    \end{array}
\end{equation*}

\begin{equation*}
      \ks \sim \rkhspace^{\smani} \qquad
\end{equation*}
\begin{equation*}
   \bigg{\{}\begin{array}{cccccc}
         \smani_\numsamp = & \{\sel_i &| 1 \leq i \leq \numsamp\}  \\
         \rkhspace^{\smani}_\numsamp = & \text{span} \{\kconfig_{\sel_i} &| \sel_i \in \smani_\numsamp\} =& \text{span}\{\lbasis_{\sel_i}^\smani &| \sel_i \in \smani_\numsamp \} \\
         \projecto_\numsamp^{\smani} :& \rkhspace^{\smani}  \to \rkhspace^{\smani}_\numsamp&\  &
    \end{array}
\end{equation*}

We also know that $\rkhspace^{\smani} = \mmapping^*(\rkhspace^\configspace$) ($\mmapping$ is assumed to be bijective)

\begin{equation*}
    ||\rkhelement||_{\rkhspace^{\smani}} := ||\rkhelement||_{\mmapping^*(\rkhspace^{\configspace})} = \text{min} \{||g||_\configspace \quad | \rkhelement = g \circ \mmapping\} 
\end{equation*}

By definition $\lbasis_{\csel_i}^\configspace(\csel_j) = \delta_{ij}$ and $\lbasis_{\sel_i}^\configspace(\sel_j) = \delta_{ij}$
\begin{equation*}
\projecto_\numsamp^\configspace g = \sum_{i=1}^m \lbasis_{\csel_i}^\configspace(\cdot)g(\csel_i) \qquad \text{for} \quad g: \configspace \to \mathbb{R}
\end{equation*}

\begin{equation*}
\projecto_\numsamp^\smani h = \sum_{i=1}^m \lbasis_{\sel_i}^\smani(\cdot)h(\sel_i) \qquad \text{for} \quad h: \smani \to \mathbb{R}
\end{equation*}

Suppose that $h(\sel) = (g\circ\mmapping)(\sel)$
\begin{equation*}
    (\projecto_\numsamp^\smani(g\circ\mmapping))(\cdot) = \sum_{i=1}^m \lbasis_{\sel_i}^S(\cdot)(g\circ\mmapping)(\sel_i) = \sum_{i=1}^m \lbasis_{\sel_i}^S(\cdot)(g)(\csel_i)
\end{equation*}

We know that the operator $\projecto_\numsamp^\smani(g\circ\mmapping)$ interpolates $g\circ\mmapping$ over $\smani_\numsamp$ and that this interpolant is unique. $\projecto_\numsamp^\configspace g$ interpolates $g$ over $\configspace_\numsamp$ which is mapped to $ (\projecto_\numsamp^\configspace g)\circ \mmapping $ which once again interpolates $g\circ\mmapping$ over $\smani_\numsamp$. Therefore the following equality holds.

\begin{equation*}
    \projecto_\numsamp^\smani(g\circ\mmapping) = (\projecto_\numsamp^\configspace g)\circ \mmapping
\end{equation*}

How to express the $\lbasis_{\sel}^\smani(\cdot)$. We know that
\begin{equation*}
    \lbasis_{\sel_i}^\smani(\cdot) = \sum_{k=1}^m\alpha_{i,k}\ks_{\sel_k}(\cdot)
\end{equation*}

\begin{equation*}
   <\ks_{\sel_j}, \lbasis_{\sel_i}^\smani(\cdot) >= \sum_{k=1}^m\alpha_{i,k}<\ks_{\sel_j},\ks_{\sel_k}>
\end{equation*}
\begin{equation*}
    \lbasis_{\sel_i}^\smani(\sel_j) = \delta_{ij} = \sum_{k=1}^m\alpha_{i,k}\ks(\sel_j,\sel_k)
\end{equation*}
Denote $K_{j,k} = k(\sel_j,\sel_k)$

\begin{equation*}
    \alpha_{i,k}=K_{j,k}^{-1}\delta_{ij} = K_{i,k}^{-1}
\end{equation*}

\begin{equation*}
    \lbasis_{\sel_i}^\smani(\cdot) = \sum_{k=1}^m K_{i,k}^{-1}\ks_{\sel_k}(\cdot)
\end{equation*}

Recall that

\begin{equation}
    E_\mu(\mmapping) = \int_\ambspace\int_\smani ||\ambel-\mmapping(\sel)||_\ambspace^2\mu(d\ambel|\sel)\mu_\smani(d\sel)
    \label{eqn:DFL}
\end{equation}

\subsection{Element-wise Homogenization}
\begin{equation*}
     (P_{\smani_\numsamp}h)(\eta,g) := \sum_{i=1}^\numsamp\frac{1}{\mu_\smani (\smani_i)}\int_{\smani_i} h(\sel,\csel)\mu_\smani(d\sel)1_{\smani_i}(\eta) 
\end{equation*} 

\begin{multline*}
     <\nu,P_{\smani_\numsamp}h>_{Y^*\times Y}\\ =\int_\smani \int_\configspace \sum_{i=1}^\numsamp\frac{1}{\mu_\smani (\smani_i)}\int_{\smani_i} h(\sel,\csel)\mu_\smani(d\sel)1_{\smani_k}(\eta)\nu(d\csel|\eta)\mu_\smani(d\eta) \\
     = \int_\smani \int_\configspace h(\sel,\csel)\sum_{i=1}^\numsamp\frac{1_\smani(\sel)}{\mu_\smani (\smani_i)}\int_{\smani_i} \mu_\smani(d\sel)1_{\smani_i}(\eta)\nu(d\csel|\eta)\mu_\smani(d\eta) \mu_\smani(d\sel) \\
     = \int_\smani \int_\configspace h(\sel,\csel)(P^*_{\smani_\numsamp} \nu)(d\csel|\eta)\mu_\smani(d\eta) \mu_\smani(d\sel)
\end{multline*}

Where
\begin{equation*}
    (P^*_{\smani_\numsamp} \nu)(d\csel|\eta) = \sum_{i=1}^\numsamp\frac{1_\smani(\sel)}{\mu_\smani (\smani_i)}\int_{\smani_i} \nu(d\csel|\eta)\mu_\smani(d\sel)1_{\smani_i}(\eta)
\end{equation*}
\subsection{Aggregation}
Let $\configspace_\numsamp \subseteq \configspace$

$\configspace_\numsamp = \{\csel_1,\dots,\csel_\numsamp\}$

Then

\begin{equation*}
    \projecto^{\configspace}_\numsamp: C(\configspace) \to \rkhspace^{\configspace}_{\numsamp} \subseteq \rkhspace^{\configspace}
\end{equation*}

Define
\begin{equation*}
     (P_{\configspace_\numsamp}h)(\eta,q) = (\projecto^{\configspace}_\numsamp h(\csel,\cdot))(\csel)= \sum_{i=1}^\numsamp  h(\eta,\csel_i)\lbasis_{\csel_i}(\csel) 
\end{equation*} 

Where $\{\lbasis_{\csel_i}(\csel) \}_{i=1}^m$ is the Lagrange basis for $\rkhspace^{\configspace_\numsamp}$.

Then we have 

\begin{multline*}
     <\nu,P_{\configspace_\numsamp}h>_{Y^*\times Y}\\ =\int_\configspace \int_\configspace \sum_{i=1}^\numsamp  h(\eta,\csel_i)\lbasis_{\csel_i}(\csel)\nu(d\csel|\eta)\mu_\smani(d\eta) \\
     = \int_\smani \int_\configspace h(\eta,\alpha)\sum_{i=1}^\numsamp\int_{\configspace} \lbasis_{\csel_i}(\csel)\nu(d\csel|\eta) \delta_{\csel_i}(\alpha)\mu_\smani(d\eta)\\
     = \int_\smani \int_\configspace h(\eta,\alpha)(P^*_{\configspace_\numsamp} \nu)(d\csel|\eta)\mu_\smani(d\eta)
\end{multline*}

So $\smani \times \configspace \subseteq \smani \times \ambspace = \sampSpace$. Let $\spartition$ be a partition of $\smani$. $\scollection = \smani_\numfunc \subseteq \smani$. $\rkhspace^{\smani}_{\numfunc} = \text{span}\{1_{\smani_i} |1 \leq i \leq n\}$. $\projecto^\smani_\numfunc :  L^2_{\mu_\smani} \to \rkhspace^{\smani}_\numfunc$ on $L^2_{\mu_\smani} = \{g:\smani \to \mathbb{R}| \; ||g||^2_{L^2_{\mu_\smani}}:= \int_{\smani}|g(\sel)|^2\mu_{\smani}(d\sel) < \infty \}$

The set $\{\frac{1_{\smani_i}}{\sqrt(\mu(\smani_i)} \}_{i=1}^n$ is $L^2_{\mu_\smani}$ and

\begin{multline*}
    \projecto_\numfunc^\smani g = \sum_{k=1}^\numfunc\int_{\ambspace} g(\ambel)\frac{1_{\smani_k}(\sel) \mu_\smani(d\sel)}{\sqrt{\mu_\smani (\smani_k)}}\frac{1_{\smani_k}(\cdot)}{\sqrt{\mu_\smani (\smani_k)}} \\= \sum_{k=1}^\numfunc\frac{\int_{\smani_k} g(\ambel)1_{\smani_k}(\sel) \mu_\smani(d\sel)}{\mu_\smani (\smani_k)}1_{\smani_k}(\cdot) 
\end{multline*}
   
Let $\cscollection =\configspace_\numsamp \subset \configspace \subseteq \ambspace$. $\{\lbasis_{\csel_i}\}_{i=1}^m$ is the Lagrange basis associated with $\{\kconfig_{\csel_i}\}_{i=1}^m$. $\rkhspace^{\configspace}_{\numsamp} = \text{span}\{\kconfig_{\csel_i}(\cdot)\} = \text{span}\{\lbasis_{\csel_i}(\cdot)\} \subseteq \rkhspace^\configspace \subseteq C(\configspace)$

\begin{equation*}
    \projecto^{\configspace}_\numsamp: \rkhspace^{\configspace} \to \rkhspace^{\configspace}_{\numsamp} 
\end{equation*}

By definition $\mmapping_\mu(\sel) = \int_\ambspace \ambel \mu(d\ambel|\sel)$. $\mmapping_\mu: \smani \to \ambspace$

\begin{equation*}
      (\projecto_\numfunc^\smani \mmapping_\mu)(\sel)= \sum_{k=1}^\numfunc\frac{1}{\mu_\smani (\smani_k)}\int_{\smani_k} \int_{\ambspace} \ambel\mu(d\ambel|\sel)1_{\smani_k}(\sel) \mu_\smani(d\sel)1_{\smani_k}(\sel)
\end{equation*} 

\begin{multline*}
    E(\mmapping) = \int_\smani \int_\ambspace ||\ambel - \mmapping(\sel) ||^2\mu(d\ambel|\sel)\mu_\smani(d\sel) \\ =\int_\smani \int_\ambspace h(\sel,\ambel)\mu(d\ambel|\sel)\mu_\smani(d\sel) = <\mu,h(\sel,\ambel)>_{Y^*\times Y}
\end{multline*}

\begin{multline*}
     E(\mmapping)  =\int_\smani \int_\ambspace h(\sel,\ambel)\mu_{\numfunc,\numsamp}(d\ambel|\sel)\mu_\smani(d\sel) \\= \int_\smani \int_\ambspace h(\sel,\ambel)(P^*_{\numfunc,\numsamp}\mu)(d\ambel|\sel)\mu_\smani(d\sel)
\end{multline*}

Recall
\begin{equation*}
    \projecto^{\smani}_\numsamp: L^2_{\mu_\smani} \to \rkhspace^{\smani}_{\numsamp} 
\end{equation*}
\begin{equation*}
   \projecto^{\configspace}_\numsamp: \rkhspace^{\ambspace} \to \rkhspace^{\configspace}_{\numsamp}
\end{equation*}

Define
\begin{equation*}
    (P^{\smani}_{\numfunc}h)(\sel,\csel)= (\projecto^{\smani}_\numsamp h(\cdot,\csel))(\sel)
\end{equation*}
\begin{equation*}
   (P^{\configspace}_{\numfunc}h)(\sel,\csel)= (\projecto^{\configspace}_\numsamp h(\sel,\cdot))(\csel)
\end{equation*}

Then
\begin{multline*}
   (P_{\numfunc,\numsamp}h)(\sel,\csel)= (P^{\smani}_{\numfunc}P^{\configspace}_{\numfunc} h)(\sel,\csel) \\
   = P^{\smani}_{\numfunc}\sum_{i=1}^\numsamp  h(\eta,\csel_i)\lbasis_{\csel_i}(\csel) \\
   = \sum_{i=1}^m \sum_{k=1}^n \int_{\smani_k}h(\eta,\csel_i)\mu_{\smani}(d\eta)\frac{1_{\smani_k}}{|\smani_k|}\lbasis_{\csel_i}(\csel)
\end{multline*}

And
\begin{multline*}
   (P_{\numfunc,\numsamp}h)(\sel,\csel)= (P^{\configspace}_{\numfunc} P^{\smani}_{\numfunc} h)(\sel,\csel) \\
   = P^{\configspace}_{\numfunc}\sum_{i=1}^\numsamp  \frac{1}{|\smani_k|}\int_{\smani_k}h(\eta,\csel_i)\lbasis_{\csel_i}(\csel)1_{\smani_k}(\sel) \\
   = \sum_{i=1}^m \sum_{k=1}^n \frac{1}{|\smani_k|}\int_{\smani_k}h(\eta,\csel_i)\mu_{\smani}(d\eta)\lbasis_{\csel_i}(\csel)
\end{multline*}

Now look at $\int h(\eta,\csel_i)\mu_{\smani}(d\csel) = \int_{\smani_k} (\csel_i-\mmapping(\eta))^2\mu_{\smani}(d\csel)$


\bibliographystyle{unsrt}
\bibliography{CDCPaperRef}

\end{document}